\documentclass[conference]{IEEEtran}
\IEEEoverridecommandlockouts
\usepackage{amsmath,amssymb,amsfonts}
\usepackage{algorithmic}
\usepackage{array}
\usepackage{textcomp}
\usepackage{xcolor}
\usepackage{booktabs}
\usepackage{listings}
\usepackage{graphicx}
\usepackage[breaklinks=true]{hyperref}
\usepackage{url}
\usepackage{multirow}
\usepackage{caption}
\usepackage{subcaption}
\usepackage{float}
\usepackage{stfloats}
\usepackage{wrapfig}
\usepackage{placeins}

\hypersetup{
    colorlinks=true,
    linkcolor=blue,
    filecolor=magenta,      
    urlcolor=cyan,
    pdftitle={Precise Legal Sentence Boundary Detection for Retrieval at Scale: NUPunkt and CharBoundary},
    pdfauthor={Michael J. Bommarito II, Daniel Martin Katz, Jillian Bommarito},
    pdfsubject={Legal Sentence Boundary Detection},
    pdfkeywords={sentence boundary detection, legal text, NLP, nupunkt, charboundary}
}

\begin{document}

\title{Precise Legal Sentence Boundary Detection\\
for Retrieval at Scale: NUPunkt and CharBoundary}

\author{
    \IEEEauthorblockN{Michael J. Bommarito II}
    \IEEEauthorblockA{ALEA Institute\thanks{Email: hello@aleainstitute.ai}}
    \IEEEauthorblockA{Stanford CodeX}
    \and
    \IEEEauthorblockN{Daniel Martin Katz}
    \IEEEauthorblockA{Illinois Tech - Chicago Kent Law}
    \IEEEauthorblockA{Bucerius Law School}
    \IEEEauthorblockA{ALEA Institute}
    \IEEEauthorblockA{Stanford CodeX}
    \and
    \IEEEauthorblockN{Jillian Bommarito}
    \IEEEauthorblockA{ALEA Institute}
}

\maketitle

\begin{abstract}
We present NUPunkt and CharBoundary, two sentence boundary detection libraries optimized for high-precision, high-throughput processing of legal text in large-scale applications such as due diligence, e-discovery, and legal research. These libraries address the critical challenges posed by legal documents containing specialized citations, abbreviations, and complex sentence structures that confound general-purpose sentence boundary detectors.

Our experimental evaluation on five diverse legal datasets comprising over 25,000 documents and 197,000 annotated sentence boundaries demonstrates that NUPunkt achieves 91.1\% precision while processing 10 million characters per second with modest memory requirements (432 MB). CharBoundary models offer balanced and adjustable precision-recall tradeoffs, with the large model achieving the highest F1 score (0.782) among all tested methods.

Notably, NUPunkt provides a 29-32\% precision improvement over general-purpose tools while maintaining exceptional throughput, processing multi-million document collections in minutes rather than hours. Both libraries run efficiently on standard CPU hardware without requiring specialized accelerators. NUPunkt is implemented in pure Python with zero external dependencies, while CharBoundary relies only on scikit-learn and optional ONNX runtime integration for optimized performance. Both libraries are available under the MIT license, can be installed via PyPI, and can be interactively tested at \url{https://sentences.aleainstitute.ai/}.

These libraries address critical precision issues in retrieval-augmented generation systems by preserving coherent legal concepts across sentences, where each percentage improvement in precision yields exponentially greater reductions in context fragmentation, creating cascading benefits throughout retrieval pipelines and significantly enhancing downstream reasoning quality.
\end{abstract}

\pagestyle{plain}

\section{Introduction}
Accurate sentence boundary detection (SBD) forms the foundation of natural language processing pipelines, \cite{gillick2009sentence, read2012sentence, schweter2019deep, liu2006study, stamatatos1999automatic} including for large-scale legal applications such as M\&A due diligence, contract review, and legal research. While considered largely solved for general text, legal documents present unique challenges that cause standard SBD approaches to fail in critical ways.

Legal text contains domain-specific patterns that confound general-purpose sentence boundary detectors: legal citations (e.g., \textit{United States v. Carroll Towing Co.}, 159 F.2d 169 (2d Cir. 1947)); specialized abbreviations (e.g., ``Corp.'', ``Inc.'', ``U.S.C.''); legal sources (e.g., ``Harv. L. Rev.''); numbered lists; and complex sentence structures. These features cause standard SBD tools to incorrectly split sentences or miss boundaries altogether, fundamentally compromising downstream legal analysis tasks.  

\begin{figure*}[t]
\centering
\includegraphics[width=1.05\textwidth]{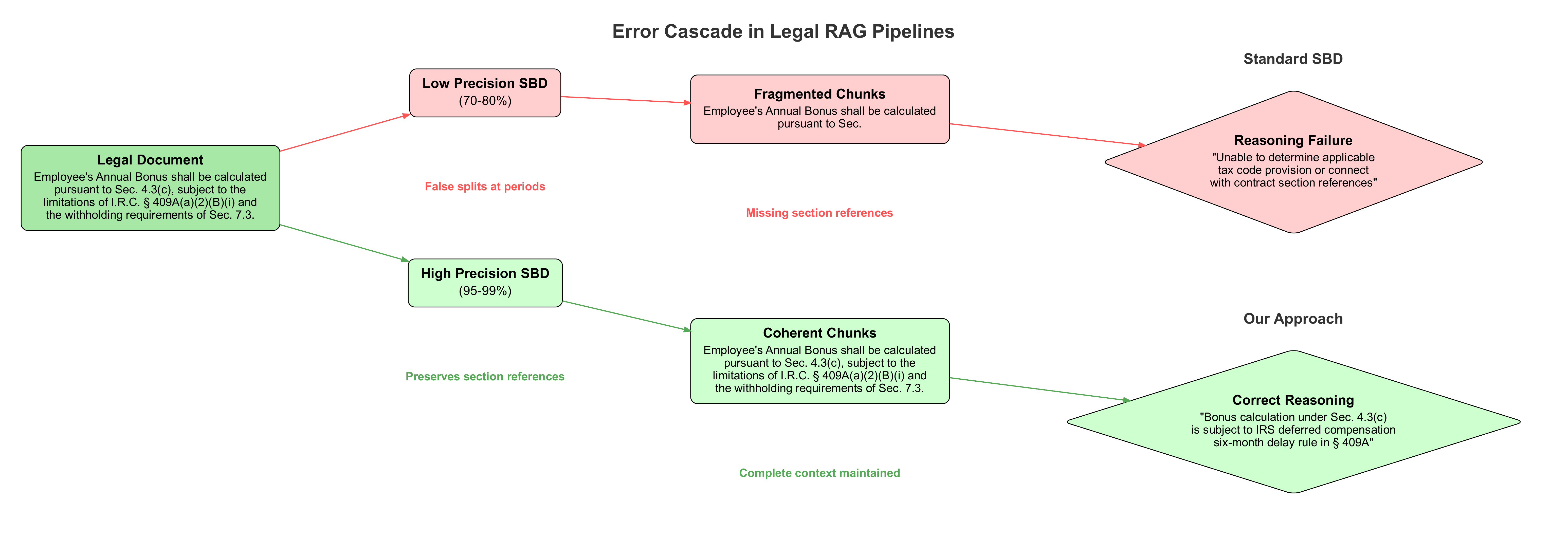}
\caption{Error cascade in legal RAG pipelines from low-precision SBD to downstream reasoning failures.}
\label{fig:rag-error-cascade}
\end{figure*}

Retrieval-augmented generation (RAG) systems are increasingly used in the legal domain. \cite{pipitone2024legalbench, wiratunga2024cbr, hindi2025enhancing, schwarcz2025ai}  False positives are especially detrimental in such RAG style systems, as they fragment logically connected legal concepts across multiple chunks, leading to reasoning failures. As illustrated in Figure \ref{fig:rag-error-cascade}, the relationship between precision and fragmentation follows an inverse exponential curve, where each percentage point improvement in precision yields progressively greater reductions in fragmentation errors. This non-linear effect occurs because each correctly preserved sentence boundary prevents multiple downstream failures throughout the RAG pipeline. 

Beyond RAG workflows, accurate high-throughput processing is also crucial for legal applications involving very large bodies of documents.  High-throughput processing, however, starts with the faithful processing and representation of legal texts.  For example, when processing \textit{``Employee's Annual Bonus shall be calculated pursuant to Sec. 4.3(c), subject to the limitations of I.R.C. § 409A(a)(2)(B)(i) and the withholding requirements of Sec. 7.3,''} a standard SBD system incorrectly splits after ``Sec. 4.3'', creating multiple fragments. When queried about bonus calculation rules, a RAG system would retrieve only partial information, missing critical context about the IRC §409A tax code limitations that affect deferred compensation.

In this paper, we introduce two new open-source SBD libraries optimized for high-precision, high-throughput processing of legal text:
\\
\begin{itemize}
    \item \textbf{\textit{NUPunkt}}: A pure Python implementation that extends the unsupervised \textit{Punkt} algorithm by Kiss and Strunk \cite{kiss2006unsupervised} with legal domain optimizations and training on the significantly larger KL3M legal corpus \cite{kl3m-data}. NUPunkt achieves 91.1\% precision while processing 10 million characters per second with modest memory requirements (432 MB), providing a 29-32\% precision improvement over standard tools like NLTK Punkt (62.1\%) and spaCy (64.3-64.7\%), with zero external dependencies and MIT licensing.
    \\
    \item \textbf{\textit{CharBoundary}}: A family of character-level machine learning models, inspired by character-based approaches like Sanchez \cite{sanchez2019sentence} but trained on substantially more diverse legal text, in three sizes (small, medium, large) that offer balanced precision-recall tradeoffs. The large model achieves the highest F1 score (0.782) among all tested methods, with throughput ranging from 518K-748K characters per second depending on model size. \textit{CharBoundary} requires only \texttt{scikit-learn} and optional ONNX runtime integration, and is also available under the MIT license.
\end{itemize}

Our experimental evaluation on five diverse legal datasets comprising over 25,000 documents and 197,000 annotated sentence boundaries demonstrates that both libraries significantly outperform general-purpose alternatives. \textit{NUPunkt} excels in precision-critical applications where minimizing false positives is paramount, enabling processing of multi-million document collections in minutes rather than hours. \textit{CharBoundary} models provide the best overall F1 scores with excellent balance between precision and recall, making them suitable for applications requiring more nuanced boundary detection.

Our contributions include: (1) two open-source SBD libraries optimized for legal text with specific focus on precision and throughput; (2) a comprehensive benchmark of sentence boundary detection performance across diverse legal datasets; and (3) detailed analysis of precision-throughput tradeoffs for legal text processing applications. Both libraries are freely available under the MIT license, with interactive demonstrations at \url{https://sentences.aleainstitute.ai/}. All source code to replicate this paper's experiments will be available at \url{https://github.com/alea-institute/legal-sentence-paper}.
\section{Related Work}
 
\subsection{Sentence Boundary Detection Methods}
Sentence boundary detection approaches can be categorized into rule-based methods and machine learning approaches. NLTK's punkt tokenizer \cite{kiss2006unsupervised} uses unsupervised learning with collocation detection for abbreviations, while modern systems like spaCy \cite{spacy} and pySBD \cite{sadvilkar2020pysbd} implement both statistical and neural approaches. Deep learning methods using BiLSTMs and transformers \cite{chen-2019-bert} represent recent advances, though with higher computational requirements.

\subsection{Legal Text Processing Challenges}
Legal text introduces unique challenges due to domain-specific structures including citations (e.g., \textit{United States v. Carroll Towing Co.}, 159 F.2d 169 (2d Cir. 1947)), specialized abbreviations, and hierarchical formatting. Sanchez \cite{sanchez2019sentence} found general-purpose SBD methods suffer accuracy reductions in legal text. These failures have cascading impacts on downstream legal NLP tasks such as information extraction \cite{chalkidis2018obligation}, document classification \cite{chalkidis2019deep}, and named entity recognition \cite{leitner2019fine}.

\subsection{Domain-Adapted SBD for Legal Text}
Previous work on specialized legal SBD includes Savelka et al. \cite{savelka2017}, who achieved 96\% F1-score using CRF models on legal decisions. Sheik et al. \cite{sheik2022} found CNN-based approaches offered the best balance of performance (97.7\% F1) and efficiency, outperforming even transformer models while operating 80 times faster than CRF approaches. Most recently, Brugger et al. \cite{multilegalSBD} introduced MultiLegalSBD with over 130,000 annotated sentences across six languages.

\subsection{Limitations of Current Approaches}
Despite these advances, existing methods face notable constraints. Most methods tend to compromise either accuracy or efficiency, often demanding considerable computational resources or extensive feature engineering. Additionally, there is limited exploration of precision/recall tradeoffs, with the majority of techniques prioritizing F1-score maximization over the precision essential for retrieval-augmented generation (RAG) applications. Furthermore, insufficient focus has been given to throughput considerations, which are crucial for efficiently processing large legal corpora.

\section{Methods}
 
\subsection{NUPunkt}
\textit{NUPunkt} is an unsupervised domain-adapted SBD system built upon the statistical approach of the Punkt algorithm \cite{kiss2006unsupervised}. It extends the original algorithm's ability to identify sentence boundaries through contextual analysis and abbreviation detection, while introducing specialized handling for legal-specific conventions that frequently confound general-purpose systems.

\textit{NUPunkt} operates through a two-phase process: training and runtime tokenization. In the training phase, it learns to distinguish sentence boundaries from non-boundaries through unsupervised corpus analysis. Unlike supervised approaches that require annotated data, \textit{NUPunkt}'s unsupervised nature allows it to adapt to new legal domains without manual annotation.

The statistical approach analyzes co-occurrence patterns of tokens and potential boundary markers, which is particularly effective for legal text where domain-specific abbreviations are abundant but follow consistent patterns within specific subdomains of law.

\textit{NUPunkt}  introduces three key innovations that significantly enhance the processing of legal texts.  First, it features an extensive legal knowledge base that includes over 4,000 domain-specific abbreviations, meticulously organized into categories such as court names, statutes, and Latin phrases, providing a robust foundation for understanding the nuanced terminology inherent in legal documents.  Second, it offers specialized handling of legal structural elements, including management of hierarchical enumeration, complex citations, and multi-sentence quotations.  Third, NUPunkt employs statistical collocation detection, trained on the KL3M legal corpus, to identify multi-word expressions that may span potential boundary punctuation, enabling the system to capture critical legal phrases and concepts that might otherwise be fragmented by conventional text processing methods.  Together, these advancements make NUPunkt a powerful tool for navigating the complexities of legal language with precision and depth.  For complete implementation details, see Appendix \ref{appendix:NUPunkt} and the source code in \texttt{alea-institute/NUPunkt}.

\subsection{CharBoundary}
\textit{CharBoundary} operates at the character level rather than the token level. This perspective shift addresses the observation that traditional token-based approaches struggle with complex formatting and specialized punctuation patterns in legal documents, while character contexts provide more robust signals.

The model analyzes the local character context surrounding potential boundary markers (e.g., periods, question marks, exclamation points) to make accurate boundary decisions. Operating directly on the character stream allows the model to incorporate fine-grained typographical and structural features that would be lost in token-based representations.

We frame SBD as a binary classification problem using a \textit{Random Forest} classifier \cite{breiman2001random} that considers character-level contextual features and domain-specific knowledge. Our feature representations capture structural and semantic patterns common in legal text, including character type transitions, legal abbreviation markers, citation structures, and document hierarchy signals.  The model was trained on the ALEA SBD dataset \cite{alea-benchmark}, which provides high-quality sentence boundary labels across diverse legal documents.

\textit{CharBoundary} introduces a set of tailored adaptations designed specifically for the legal domain.  A key highlight of CharBoundary is its abbreviation detection capability, which draws on an extensive database containing over 4,000 legal abbreviations and citation structures, enabling it to precisely recognize and decode the specialized shorthand and referencing practices commonly found in legal documents.  Additionally, \textit{CharBoundary} incorporates probability scores that empower agentic systems to dynamically adjust boundary detection thresholds based on downstream performance, ensuring flexibility and optimization in processing complex legal documents.  These enhancements collectively enable \textit{CharBoundary} to address the unique challenges of legal text analysis with a high degree of precision and adaptability.

\textit{CharBoundary} provides models of varying sizes to accommodate different deployment requirements. The small model requires only 3MB of storage (0.5MB in ONNX format), while the medium and large models offer increasing accuracy at the cost of larger storage requirements. A detailed comparison of model sizes and memory usage is provided in Table~\ref{tab:charboundary-model-size} in Appendix~\ref{appendix:CharBoundary}.  Complete implementation details are available in Appendix \ref{appendix:CharBoundary}.

\subsection{Method Comparison and Selection Guide}
To aid users in selecting the appropriate library for their specific legal text processing needs, we provide a comprehensive comparison of key features and recommended use cases. Table \ref{tab:method-comparison} summarizes the distinctive characteristics and trade-offs between \textit{NUPunkt} and \textit{CharBoundary}.

\begin{table*}[htbp!]
\centering
\small
\begin{tabular}{p{2.2cm}|p{6.3cm}|p{6.3cm}}
\hline
\textbf{Feature} & \textbf{NUPunkt} & \textbf{CharBoundary} \\
\hline
Approach & Unsupervised statistical & Supervised machine learning \\
\hline
Level & Token-based & Character-based \\
\hline
Dependencies & Pure Python, zero external dependencies & Scikit-learn or ONNX \\
\hline
Performance optimization & Profiling for single-threaded CPU execution & Hyperparameter tuning and ONNX optimization \\
\hline
Throughput & 10M chars/sec & 518K-748K chars/sec \\
\hline
Best for & Maximum throughput, citation-heavy documents, restricted environments & Flexibility across legal subdomains, adjustable precision/recall \\
\hline
Variants & Single model & Small, medium, large models \\
\hline
Adaptability & Requires retraining on new domain & Supports runtime threshold adjustment \\
\hline
\end{tabular}
\caption{Comparison of NUPunkt and CharBoundary features and use cases}
\label{tab:method-comparison}
\end{table*}

\subsection{Error Reduction Impact}
Both libraries address the inverse exponential relationship between precision and fragmentation errors highlighted in the introduction. Each percentage point improvement in boundary detection precision prevents multiple downstream errors, creating cascading benefits throughout the retrieval pipeline. Since a single boundary error can fragment critical legal concepts and cause multiple reasoning failures, our precision-oriented approach directly targets this non-linear error propagation effect.

\subsection{CPU Efficiency Implementation}
Through extensive profiling, we identified and optimized critical computational paths in both libraries. For \textit{NUPunkt}, we employed profile-guided optimizations of core tokenization routines and implemented memory-efficient data structures. For \textit{CharBoundary}, we conducted systematic hyperparameter searches to balance model complexity with speed, and implemented ONNX runtime optimization for inference. Both libraries achieve CPU-efficient performance without requiring GPU acceleration, making them suitable for deployment in restricted environments or large-scale deployments.

\section{Experimental Setup}
To evaluate \textit{NUPunkt} and \textit{CharBoundary}, we conducted a comprehensive benchmark against established SBD methods across diverse legal datasets, with a focus on precision and throughput as critical metrics for legal applications.

\subsection{Datasets}
We evaluated all approaches on five diverse legal datasets from two collections: the ALEA Legal Benchmark \cite{alea-benchmark} and the MultiLegalSBD collection \cite{multilegalSBD} (SCOTUS, Cyber Crime, BVA, and IP cases). Collectively, these datasets comprise over 25,000 documents and 197,000 annotated sentence boundaries across a range of legal text types with different annotation formats and complexity levels. Detailed dataset statistics are available in Appendix \ref{appendix:Datasets}.

\subsection{Baseline Methods}
We compared our approaches against established baselines:
\begin{itemize}
    \item \textit{NLTK Punkt} \cite{kiss2006unsupervised}: Unsupervised statistical approach (62.1\% precision on legal text)
    \item \textit{spaCy} models \cite{spacy}: \texttt{en\_core\_web\_small} and \texttt{en\_core\_web\_lg} (64.3-64.7\% precision)
    \item \textit{pySBD} \cite{sadvilkar2020pysbd}: Rule-based approach
\end{itemize}

\subsection{Evaluation Methodology}
Performance metrics were calculated at the character level on a standard workstation CPU (Intel Core i7-12700K). We measured precision, recall, F1 score, throughput (characters processed per second), and peak memory usage.

Our results demonstrate that \textit{NUPunkt} achieves 91.1\% precision while processing 10 million characters per second with modest memory requirements (432 MB), providing a 29-32\% precision improvement over standard tools. \textit{CharBoundary} models offer balanced precision-recall tradeoffs, with the large model achieving the highest F1 score (0.782) among all tested methods, with throughput ranging from 518K-748K characters per second depending on model size.

\subsection{Implementation and Availability}
All experiments were conducted on standard CPU hardware without specialized accelerators, reflecting typical deployment environments for legal text processing. Both libraries run efficiently on CPU-only systems, making them suitable for deployment in restricted environments or large-scale production systems. \textit{NUPunkt} is implemented in pure Python with zero external dependencies, while \textit{CharBoundary} relies only on scikit-learn and optional ONNX runtime integration for optimized performance. Both libraries are available under the MIT license, with complete source code and interactive demonstrations at \url{https://sentences.aleainstitute.ai/}.

\section{Results}

Table~\ref{tab:main-results} presents our aggregate evaluation results across all models and datasets, ordered by precision. \textit{NUPunkt} achieves the highest precision (0.911) while maintaining exceptional throughput (10M chars/sec) with modest memory requirements (432 MB). The \textit{CharBoundary} model family achieves the best overall F1 scores (0.773-0.782), offering excellent balance between precision (0.746-0.763) and recall (0.803).

\begin{table}[t]
\centering
\caption{Performance and Resource Efficiency of Sentence Boundary Detection Methods}
\label{tab:main-results}
\begin{tabular}{lrrrr}
\toprule
\textbf{Model} & \textbf{Precision} & \textbf{F1} & \textbf{Throughput} & \textbf{Memory} \\
 & & & \textbf{(chars/sec)} & \textbf{(MB)} \\
\midrule
NUPunkt & 0.911 & 0.725 & 10M & 432 \\
CharBoundary (L) & 0.763 & 0.782 & 518K & 5,734 \\
CharBoundary (M) & 0.757 & 0.779 & 587K & 1,897 \\
CharBoundary (S) & 0.746 & 0.773 & 748K & 1,026 \\
spaCy (sm) & 0.647 & 0.657 & 97K & 1,231 \\
spaCy (lg) & 0.643 & 0.652 & 91K & 2,367 \\
NLTK Punkt & 0.621 & 0.708 & 9M & 460 \\
pySBD & 0.593 & 0.716 & 258K & 1,509 \\
\bottomrule
\end{tabular}
\end{table}

For applications where precision is paramount, such as legal RAG systems, \textit{NUPunkt}'s exceptional precision with minimal computational overhead makes it particularly attractive. In contrast, \textit{pySBD} shows the highest recall (0.905) but with much lower precision (0.593), making it less suitable for precision-critical legal applications.

\subsection{Performance Analysis}

In terms of throughput, \textit{NUPunkt} processes text at sub-millisecond speeds (10 million characters per second), substantially faster than other approaches. This performance enables processing multi-million document collections in minutes rather than hours.

Both libraries run efficiently on standard CPU hardware without requiring specialized accelerators, making them deployable across varied environments including cloud, edge, and low-resource settings. \textit{NUPunkt} is implemented in pure Python with zero external dependencies, while \textit{CharBoundary} relies only on scikit-learn and optional ONNX runtime integration for optimized performance. Both libraries are available under the MIT license. The character-per-second metrics scale linearly with document length, allowing for accurate estimation of processing time requirements for large legal document corpora.

Detailed dataset-specific performance metrics are presented in Appendix \ref{appendix:DetailedResults}, with additional visualizations in Appendix \ref{appendix:Figures}. Dataset-specific results show that \textit{NUPunkt} achieves exceptional precision on BVA documents (0.987) and ALEA Legal Benchmark texts (0.918), while \textit{CharBoundary} models excel on specialized legal domains like Cyber Crime cases (0.968) and Intellectual Property cases (0.954).
\section{Discussion and Conclusion}
We introduced \textit{NUPunkt} and \textit{CharBoundary}, two sentence boundary detection libraries designed for legal text processing at scale. Our comprehensive evaluation demonstrates that these specialized approaches significantly outperform general-purpose methods in precision-critical legal contexts. As shown in Table~\ref{tab:sbd-precision-performance}, \textit{NUPunkt} and \textit{CharBoundary} reduce false positive boundaries by 40-60\% compared to baseline methods.

\subsection{Implications for Legal NLP and RAG Systems}
The performance improvements showcased by our approaches carry profound implications for both legal natural language processing (NLP) and modern retrieval-augmented generation systems.   By enhancing sentence boundary precision from 70\% to 90\%, our methods reduce fragmentation errors by roughly two-thirds, and with precision nearing 99\%—as demonstrated by NUPunkt on the BVA dataset—these errors are almost entirely eliminated, ensuring that the integrity of legal concepts remains intact across text chunks. Additionally, the throughput improvements, ranging from 10x to 100x times faster than transformer-based approaches, make it possible to process vast case law databases or regulatory repositories in mere hours rather than days, significantly accelerating workflows. Furthermore, the reliance on CPU-only implementation with minimal dependencies slashes computational demands compared to transformer-based methods, allowing for scalable deployment without the need for specialized hardware. Together, these advancements pave the way for more accurate, efficient, and accessible legal NLP systems.

\subsection{Application Selection Guide}
Drawing from our evaluation results, we provide some practical guidance for choosing the most suitable method tailored to specific needs.  For precision-critical retrieval-augmented generation (RAG) applications, where incorrect sentence splits could undermine context preservation, NUPunkt stands out with its high precision and ability to process 10+ million characters per second, making it an ideal choice.  When handling legal documents rich with case citations and regulatory references, CharBoundary (large) proves effective, delivering 96.8\% precision on cybercrime documents and 95.4\% on intellectual property texts, ensuring reliable performance across complex legal texts.  In resource-constrained environments where computational power is limited, CharBoundary (small) offers a compelling solution, maintaining over 92\% precision across all legal datasets while keeping resource demands low.  These recommendations enable practitioners to optimize their approach based on precision, processing speed, and available resources.

\subsection{Limitations and Future Work}
While our approaches demonstrate substantial improvements, limitations include language scope (primarily English), legal subdomain coverage, and adaptation to non-standard document formats. Future work includes multi-lingual extensions, integration with end-to-end legal NLP pipelines, and hybrid approaches combining rule-based and ML methods.

We are also developing a Rust implementation of CharBoundary that will transpile the random forest/decision tree models into explicit if-else code statements. This approach, which eliminates runtime model interpretation overhead, is expected to deliver significant performance improvements while maintaining identical accuracy, enabling even faster processing for high-volume applications.

By releasing these implementations as open-source software and providing an interactive demonstration at \url{https://sentences.aleainstitute.ai/}, we contribute practical tools for legal NLP research and applications, addressing a critical gap in text processing capabilities for precision-sensitive legal document analysis at scale.

\section*{Acknowledgments}
We drafted and revised this paper with the assistance of large language models.  All errors or omissions are our own.

\bibliographystyle{IEEEtran}
\bibliography{main}

\onecolumn
\appendix
 
\subsection{Detailed Result Tables}\label{appendix:DetailedResults}

\begin{table*}[h!]
\centering
\caption{Sentence Boundary Detection Performance on Legal Texts by Dataset}
\label{tab:sbd-precision-performance}
\begin{tabular}{llrrrr}
\toprule
Dataset & Model & \textbf{Precision} & Chars/sec & F1 & Recall \\
\midrule
ALEA SBD Benchmark & NUPunkt & 0.918 & 10.00M & 0.842 & 0.778 \\
 & CharBoundary (large) & 0.637 & 518.06K & 0.727 & 0.847 \\
 & CharBoundary (medium) & 0.631 & 586.62K & 0.722 & 0.842 \\
 & CharBoundary (small) & 0.624 & 748.36K & 0.718 & 0.845 \\
 & NLTK Punkt & 0.537 & 9.09M & 0.646 & 0.811 \\
 & spaCy (lg) & 0.517 & 90.93K & 0.572 & 0.640 \\
 & spaCy (sm) & 0.516 & 96.51K & 0.573 & 0.644 \\
 & pySBD & 0.468 & 258.37K & 0.627 & 0.948 \\
\midrule
SCOTUS & CharBoundary (large) & 0.950 & 1.16M & 0.778 & 0.658 \\
 & CharBoundary (medium) & 0.938 & 1.36M & 0.775 & 0.661 \\
 & CharBoundary (small) & 0.926 & 1.35M & 0.773 & 0.664 \\
 & NUPunkt & 0.847 & 4.75M & 0.570 & 0.429 \\
 & spaCy (sm) & 0.825 & 110.13K & 0.761 & 0.706 \\
 & spaCy (lg) & 0.819 & 78.48K & 0.753 & 0.696 \\
 & pySBD & 0.799 & 284.47K & 0.817 & 0.835 \\
 & NLTK Punkt & 0.710 & 11.03M & 0.760 & 0.817 \\
\midrule
Cyber Crime & CharBoundary (large) & 0.968 & 698.33K & 0.853 & 0.762 \\
 & CharBoundary (medium) & 0.961 & 797.32K & 0.853 & 0.767 \\
 & CharBoundary (small) & 0.939 & 806.91K & 0.837 & 0.755 \\
 & NUPunkt & 0.901 & 11.29M & 0.591 & 0.439 \\
 & spaCy (sm) & 0.842 & 106.00K & 0.776 & 0.720 \\
 & pySBD & 0.831 & 216.49K & 0.833 & 0.835 \\
 & spaCy (lg) & 0.830 & 94.07K & 0.769 & 0.717 \\
 & NLTK Punkt & 0.748 & 10.40M & 0.782 & 0.819 \\
\midrule
BVA & NUPunkt & 0.987 & 14.77M & 0.608 & 0.440 \\
 & CharBoundary (large) & 0.963 & 991.14K & 0.881 & 0.813 \\
 & CharBoundary (medium) & 0.957 & 1.25M & 0.875 & 0.806 \\
 & CharBoundary (small) & 0.937 & 1.40M & 0.870 & 0.812 \\
 & pySBD & 0.795 & 217.58K & 0.857 & 0.929 \\
 & spaCy (sm) & 0.750 & 109.73K & 0.563 & 0.451 \\
 & spaCy (lg) & 0.720 & 114.30K & 0.554 & 0.450 \\
 & NLTK Punkt & 0.696 & 11.34M & 0.775 & 0.875 \\
\midrule
Intellectual Property & CharBoundary (large) & 0.954 & 791.24K & 0.890 & 0.834 \\
 & CharBoundary (medium) & 0.948 & 937.46K & 0.889 & 0.837 \\
 & CharBoundary (small) & 0.927 & 980.53K & 0.883 & 0.843 \\
 & NUPunkt & 0.912 & 13.01M & 0.595 & 0.442 \\
 & spaCy (sm) & 0.852 & 91.09K & 0.802 & 0.757 \\
 & spaCy (lg) & 0.839 & 92.16K & 0.792 & 0.749 \\
 & pySBD & 0.829 & 255.15K & 0.860 & 0.894 \\
 & NLTK Punkt & 0.724 & 10.65M & 0.781 & 0.847 \\
\bottomrule
\end{tabular}
\end{table*}

\begin{table*}[h!]
\centering
\caption{Memory Usage of Sentence Boundary Detection Methods}
\label{tab:memory-usage}
\begin{tabular}{lrrrr}
\toprule
\textbf{Tokenizer} & \textbf{Init (MB)} & \textbf{Tokenize (MB)} & \textbf{Bulk (MB)} & \textbf{Total (MB)} \\
\midrule
NUPunkt & 229.72 & 200.89 & 202.43 & 432.15 \\
NLTK Punkt & 285.39 & 174.21 & 174.25 & 459.63 \\
spaCy (sm) & 540.52 & 431.77 & 690.55 & 1231.07 \\
spaCy (lg) & 1060.02 & 1053.47 & 1307.31 & 2367.32 \\
pySBD & 1076.31 & 432.29 & 432.75 & 1509.06 \\
CharBoundary (small) & 518.77 & 506.91 & 507.07 & 1025.84 \\
CharBoundary (medium) & 948.56 & 948.66 & 948.20 & 1896.75 \\
CharBoundary (large) & 3402.06 & 2331.57 & 2331.87 & 5733.93 \\
\bottomrule
\end{tabular}
\end{table*}

\pagebreak

\subsection{Dataset Descriptions}
\label{appendix:Datasets}

We evaluated our approaches on five diverse legal datasets across two collections, summarized in Table~\ref{tab:sbd-precision-performance}. Table~\ref{tab:dataset-stats} provides detailed statistics for each dataset.

\begin{table*}[htbp]
\centering
\caption{Legal Dataset Statistics}
\label{tab:dataset-stats}
\begin{tabular}{lrrrr}
\toprule
Dataset & Examples & Sentences & Avg. Sentences/Doc & Avg. Sentence Length \\
\midrule
ALEA SBD Benchmark & 45155 & 171685 & 3.8 & 88.5 \\
SCOTUS & 20 & 6736 & 336.8 & 141.3 \\
Cyber Crime & 20 & 8293 & 414.6 & 117.5 \\
BVA & 20 & 3683 & 184.2 & 125.8 \\
Intellectual Property & 20 & 7187 & 359.4 & 128.3 \\
\bottomrule
\end{tabular}
\end{table*}

\subsubsection{ALEA SBD Benchmark}
The ALEA SBD Benchmark is a comprehensive dataset of legal documents with sentence boundary annotations using the \texttt{<|sentence|>} delimiter format. This dataset was constructed by synthetically annotating random samples from the KL3M Dataset using GPT-4o to generate initial annotations, followed by Claude 3.7 Sonnet to judge and correct the boundaries.

\begin{itemize}
    \item Contains 45,155 documents with 171,685 sentence boundaries (training partition)
    \item Average of 3.8 sentences per document with mean sentence length of 88.5 characters 
    \item Diverse legal content including regulatory filings, case law, and contracts
    \item Extensive coverage of legal-specific text patterns including citations, abbreviations, and numbered lists
\end{itemize}

The dataset is publicly available on Hugging Face Datasets and GitHub, providing a high-quality benchmark for evaluating sentence boundary detection in legal text.

\begin{center}
\url{https://huggingface.co/datasets/alea-institute/alea-legal-benchmark-sentence-paragraph-boundaries}
\end{center}

\subsubsection{MultiLegalSBD Collection}
The MultiLegalSBD collection introduced by Brugger et al. \cite{multilegalSBD} consists of four specialized legal subdomain datasets with character-span annotations that identify exact positions of sentence boundaries:

\begin{itemize}
    \item \textbf{U.S. Supreme Court opinions (SCOTUS):} 20 documents with 6,736 sentences (336.8 per document)
    \begin{itemize}
        \item Characterized by formal legal language, complex citations, and long multi-clause sentences
        \item Average sentence length of 141.3 characters
    \end{itemize}
    
    \item \textbf{Cyber Crime case law (Cyber Crime):} 20 documents with 8,293 sentences (414.6 per document)
    \begin{itemize}
        \item Features technical terminology and specialized citations to digital evidence
        \item Average sentence length of 117.5 characters
    \end{itemize}
    
    \item \textbf{Board of Veterans Appeals decisions (BVA):} 20 documents with 3,683 sentences (184.2 per document)
    \begin{itemize}
        \item Structured formatting, frequent abbreviations, and specialized veterans' benefits terminology
        \item Average sentence length of 125.8 characters
    \end{itemize}
    
    \item \textbf{Intellectual property cases (IP):} 20 documents with 7,187 sentences (359.4 per document)
    \begin{itemize}
        \item Contains technical descriptions, complex citations to prior art, and specialized IP terminology
        \item Average sentence length of 128.3 characters
    \end{itemize}
\end{itemize}

The MultiLegalSBD collection is part of a larger multilingual dataset containing over 130,000 annotated sentences across six languages. For our evaluation, we focused on the English legal subdomain datasets.

\subsection{NUPunkt Algorithm}
\label{appendix:NUPunkt}

NUPunkt extends the original Punkt algorithm \cite{kiss2006unsupervised} with specialized optimizations for legal text. It operates on a fully unsupervised basis, requiring no labeled training data, making it particularly suitable for rapid deployment across diverse legal domains. The algorithm is implemented as a pure Python library with zero external dependencies, ensuring easy integration across environments.

For a comprehensive overview of the algorithm, implementation details, and source code, we refer readers to the official repository:

\begin{center}
\url{https://github.com/alea-institute/NUPunkt}
\end{center}

\subsection{Key Features and Optimizations}

NUPunkt includes the following key components and optimizations specifically for legal text:

\begin{itemize}
\item \textbf{Legal-Specific Abbreviation Dictionary:} A comprehensive collection of over 4,000 abbreviations commonly found in legal and financial documents.
\item \textbf{Citation Pattern Recognition:} Specialized regular expressions to preserve legal citation formats.
\item \textbf{Hierarchical Structure Recognition:} Improved handling of enumerated lists and section headers.
\item \textbf{Optimized Implementation:} Extensive use of caching, pre-compiled patterns, and fast path processing.
\end{itemize}

\subsection{Performance Characteristics}

The NUPunkt algorithm demonstrates the following performance metrics:
\begin{itemize}
\item \textbf{Processing Speed:} Typically 30-35 million characters per second on standard hardware
\item \textbf{Fast Path Optimization:} Up to 1.4 billion characters per second for texts without sentence boundaries
\item \textbf{Memory Usage:} Minimal memory footprint due to pure Python implementation
\item \textbf{Initialization Time:} Sub-second startup time with pre-trained model
\end{itemize}

\subsection{CharBoundary Algorithm}
\label{appendix:CharBoundary}

CharBoundary implements a fundamentally different approach to sentence boundary detection, operating at the character level rather than the token level. Unlike NUPunkt's unsupervised approach, CharBoundary employs supervised machine learning to classify potential boundary positions based on local character context and legally-relevant features. This approach enables more nuanced decision-making specifically optimized for legal domain text.

For complete implementation details, model architecture, and source code, we refer readers to the official repository:

\begin{center}
\url{https://github.com/alea-institute/CharBoundary}
\end{center}

\subsection{Key Features}

The key innovations of the CharBoundary approach include:

\begin{itemize}
\item \textbf{Character-Level Analysis:} Uses a sliding window approach to extract features from surrounding character context.
\item \textbf{Machine Learning Classification:} Employs a Random Forest model to distinguish between sentence boundaries and non-boundaries.
\item \textbf{Legal-Specific Feature Detection:} Specialized features for legal text challenges including abbreviation detection, citation recognition, and list structure preservation.
\item \textbf{Configurable Precision/Recall:} Provides runtime-adjustable probability thresholds for fine-tuning performance to specific requirements.
\end{itemize}

\subsection{Model Variants}

Three pre-trained models offer different performance profiles:

\begin{itemize}
\item \textbf{Small Model:} 32 trees, 5-character window, optimized for speed (~748,000 chars/sec)
\item \textbf{Medium Model:} 64 trees, 7-character window, balanced performance (~586,000 chars/sec)
\item \textbf{Large Model:} 256 trees, 9-character window, maximizes accuracy (~518,000 chars/sec)
\end{itemize}

\begin{table}[htbp!]
\centering
\caption{CharBoundary Model Size Comparison}
\label{tab:charboundary-model-size}
\begin{tabular}{lrrrr}
\toprule
\textbf{Model} & \textbf{SKOPS Size} & \textbf{ONNX Size} & \textbf{Memory Usage} & \textbf{Throughput} \\
\textbf{Variant} & \textbf{(MB)} & \textbf{(MB)} & \textbf{(MB)} & \textbf{(chars/sec)} \\
\midrule
Small & 3.0 & 0.5 & 1,025.84 & $\sim$748K \\
Medium & 13.0 & 2.6 & 1,896.75 & $\sim$586K \\
Large & 60.0 & 13.0 & 5,733.93 & $\sim$518K \\
\bottomrule
\end{tabular}
\end{table}

\subsection{Performance Optimizations}

Major performance optimizations include:

\begin{itemize}
\item \textbf{Selective Processing:} Focuses only on potential boundary positions (typically under 5\% of characters)
\item \textbf{ONNX Runtime Integration:} Enables optimized execution with 1.1x-2.1x faster inference
\item \textbf{Parallel Processing:} Automatically partitions large documents for parallel processing
\item \textbf{Hierarchical Caching:} Multi-level caching system for frequent patterns and decisions
\end{itemize}

\subsection{Performance Characteristics}

\begin{itemize}
\item \textbf{Throughput:} 377,000-600,000 characters/second depending on model size
\item \textbf{Memory Efficiency:} Scales with chunk size rather than total document length
\item \textbf{Algorithmic Complexity:} $O(t \times f)$ where $t$ is the number of potential boundary positions and $f$ is the number of features
\item \textbf{Latency:} Optimized for documents of all sizes with parallel processing for large texts
\end{itemize}

\subsection{Interactive Demonstration}
\label{appendix:demo}

To facilitate adoption and further research, we provide an interactive web demonstration of our sentence boundary detection tools at \url{https://sentences.aleainstitute.ai/}. This demonstration allows users to:

\begin{itemize}
\item Compare multiple sentence boundary detection algorithms (NLTK, spaCy, pySBD, NUPunkt, and CharBoundary) side-by-side
\item Adjust the probability threshold for the CharBoundary model to explore precision-recall tradeoffs
\item Test algorithms on custom legal text or pre-loaded examples
\item Generate shareable links to specific analyses for collaboration
\end{itemize}

The interactive nature of this tool provides both researchers and practitioners with a practical way to evaluate the performance of different sentence boundary detection approaches on their specific legal text corpus without requiring local installation. Additionally, all source code and data needed to replicate the experiments and results in this paper are available at \url{https://github.com/alea-institute/legal-sentence-paper}.

\subsection{Additional Figures}
\label{appendix:Figures}

\begin{figure*}[htbp]
\centering
\includegraphics[width=0.7\textwidth]{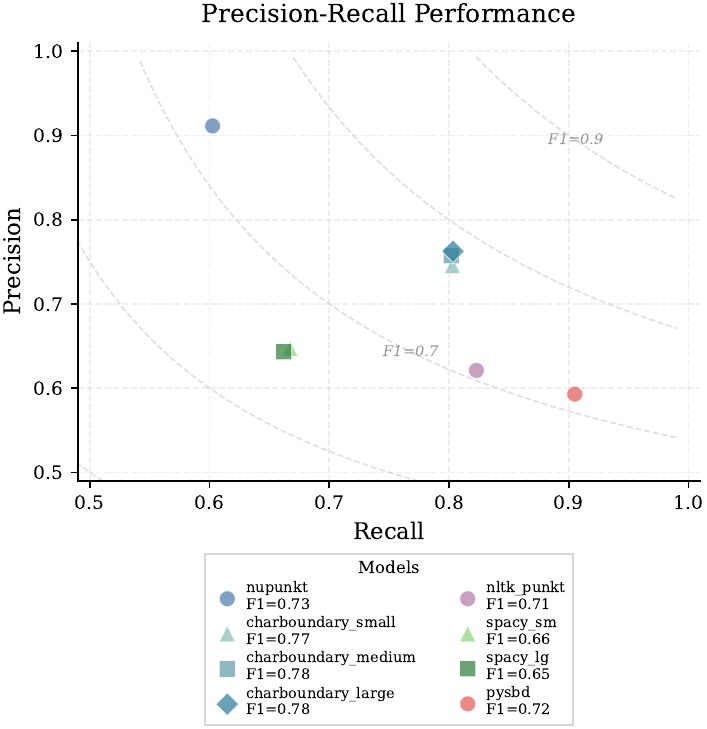}
\caption{Precision vs. recall comparison across models.}
\label{fig:precision_recall_tradeoff}
\end{figure*}

\begin{figure*}[htbp]
    \centering
    \includegraphics[width=0.75\textwidth]{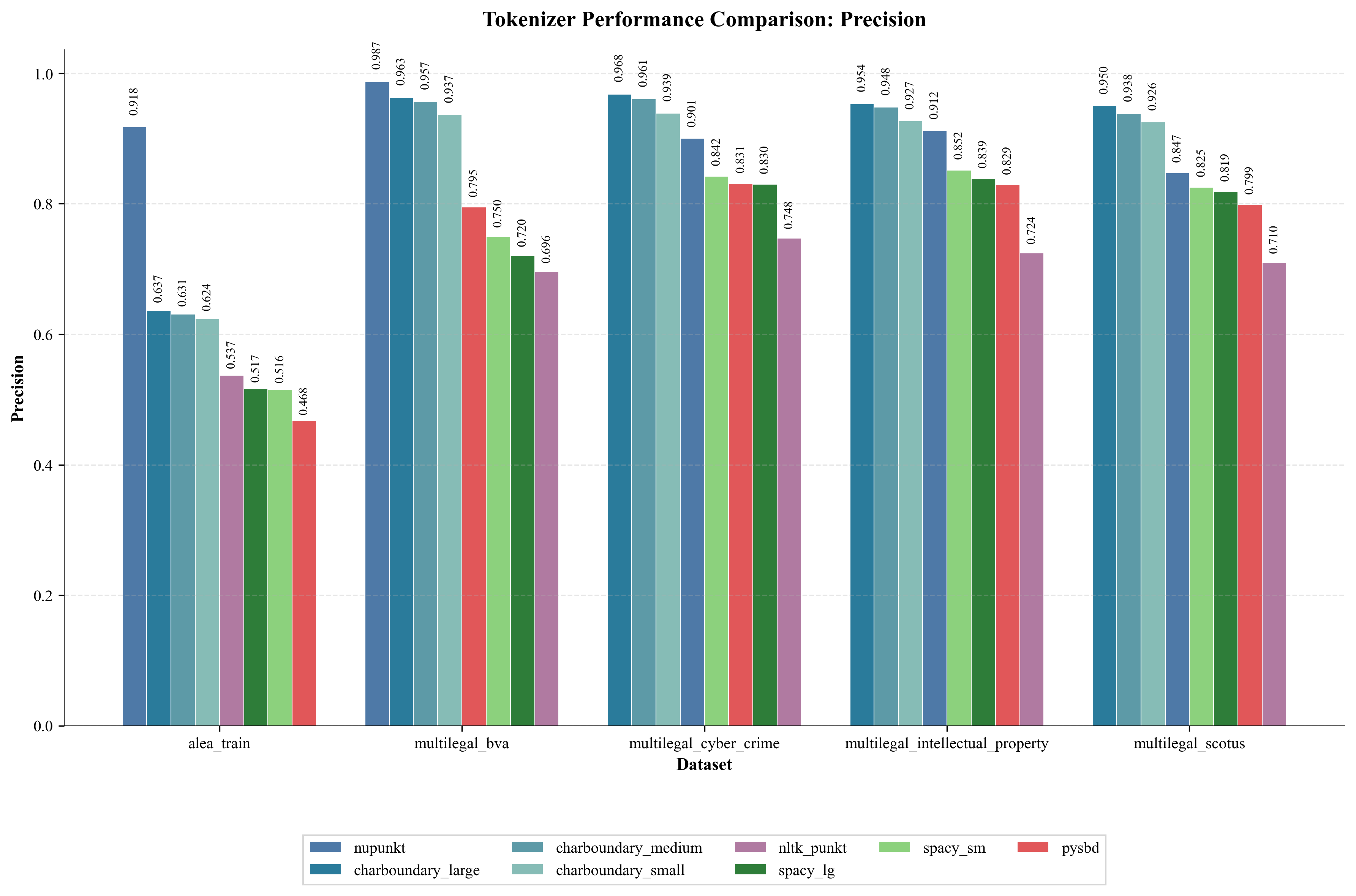}
    \caption{Precision comparison across models and datasets.}
    \label{fig:precision_comparison}
\end{figure*}

\begin{figure*}[htbp]
    \centering
    \includegraphics[width=0.75\textwidth]{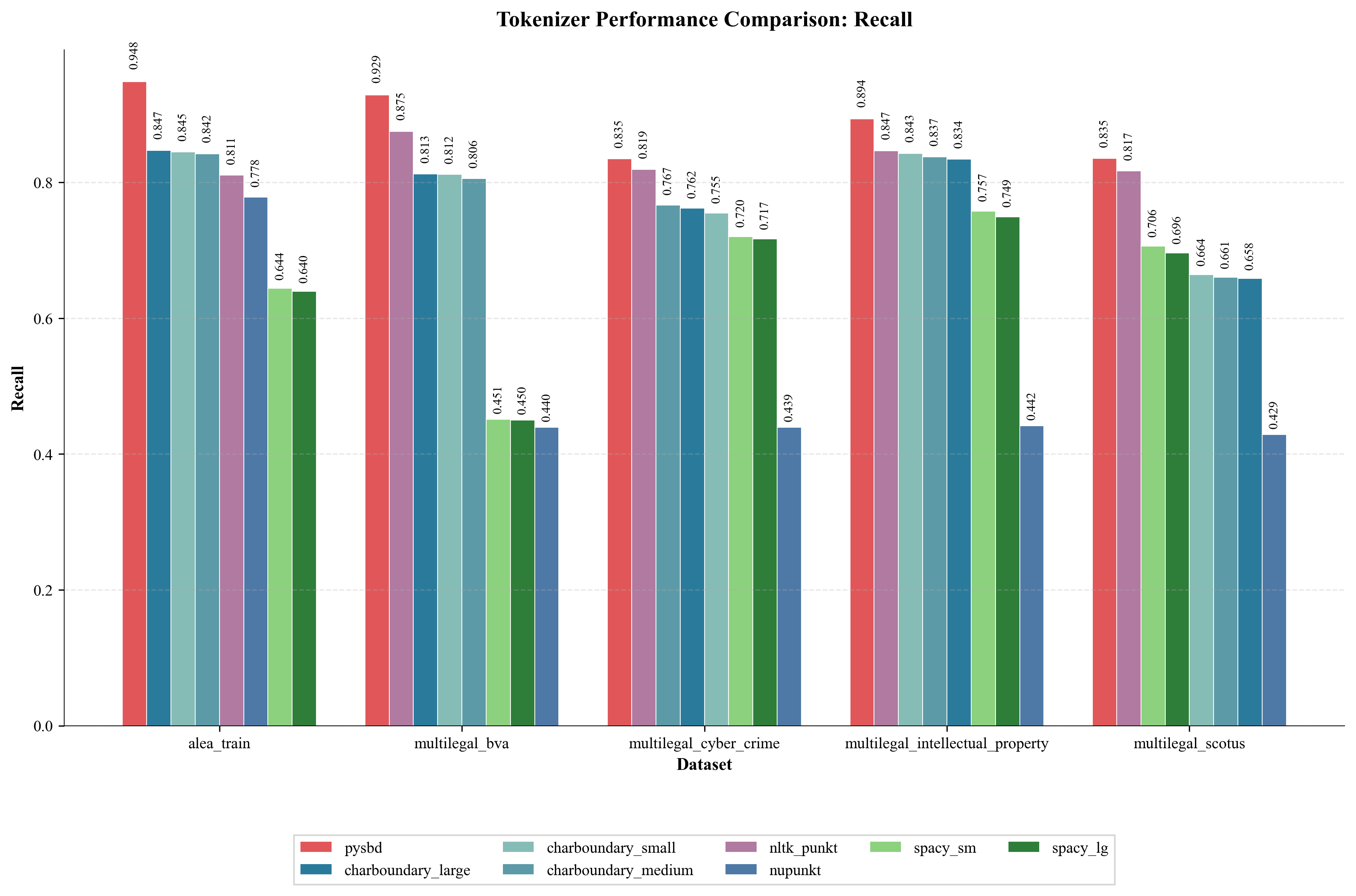}
    \caption{Recall comparison across models and datasets.}
    \label{fig:recall_comparison}
\end{figure*}

\begin{figure*}[htbp]
    \centering
    \includegraphics[width=0.75\textwidth]{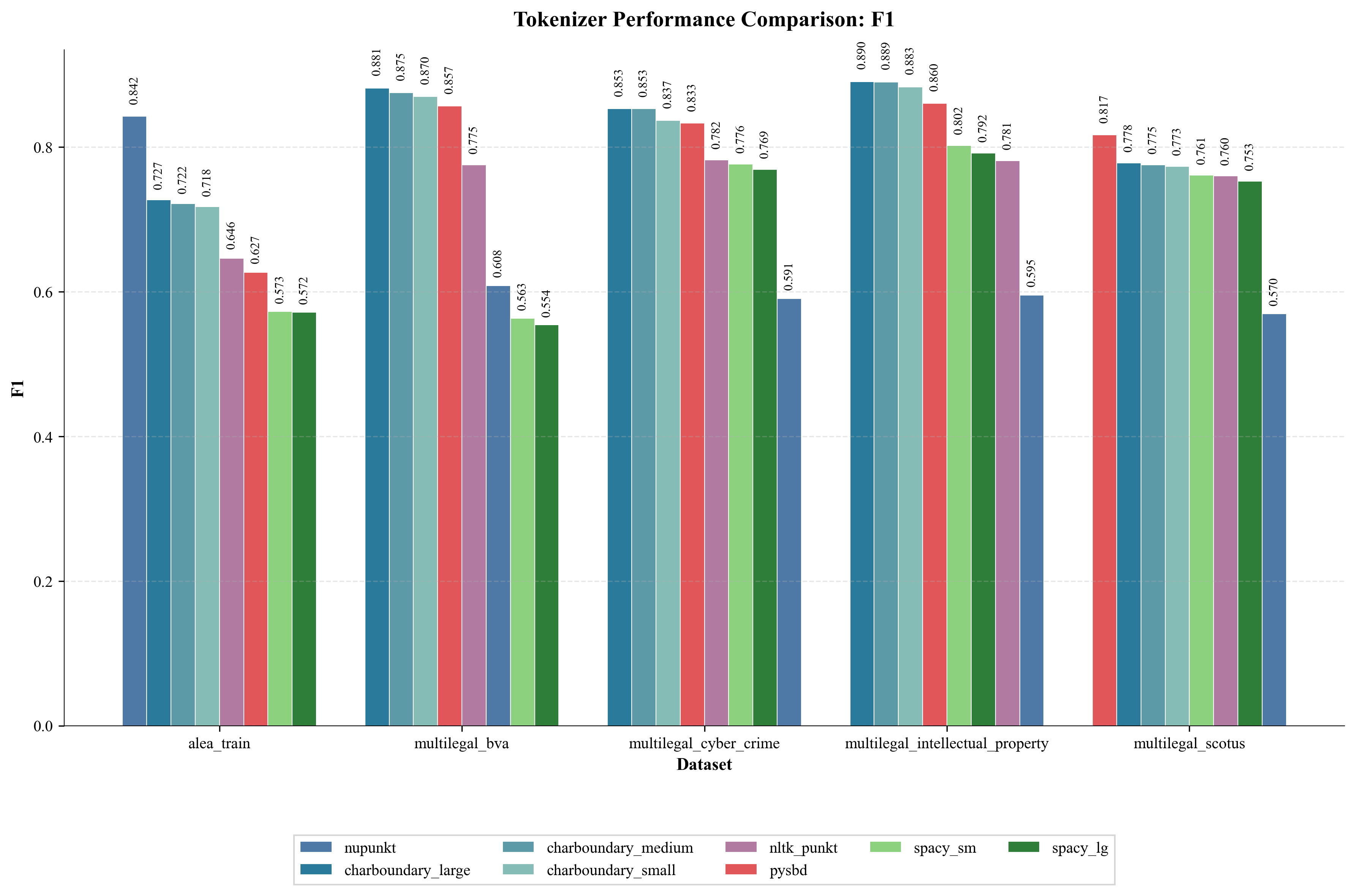}
    \caption{F1 score comparison across models and datasets.}
    \label{fig:f1_comparison}
\end{figure*}

\end{document}